\documentclass[letterpaper, 10 pt, conference]{ieeeconf}  % Comment this line out if you need a4paper

\IEEEoverridecommandlockouts                              % This command is only needed if 
\usepackage{graphicx} % for pdf, bitmapped graphics files
\usepackage{epsfig} % for postscript graphics files
\usepackage{amsmath} 
\usepackage{amssymb}  
\usepackage{soul,color} 
\usepackage{mathtools}
\usepackage{booktabs}
\usepackage{multirow}
\usepackage{enumerate}
\usepackage{balance}
\usepackage{gensymb}
\usepackage{tabulary}
\usepackage{dblfloatfix}
\usepackage{layouts}
\usepackage{algorithm}
\usepackage[noend]{algorithmic}
\usepackage{cite}
\usepackage{bm}
\usepackage{hyperref}
\usepackage{svg}
\usepackage{cite}
\usepackage{subcaption}
\usepackage{colortbl}
\usepackage{float}
\usepackage{svg}

\def\baselinestretch{1}\selectfont

\newcolumntype{?}{!{\vrule width 2pt}}
\definecolor{lightgreen}{rgb}{0.8,1,0.8}

\usepackage{caption}
\graphicspath{{../figs/}}

\makeatletter
\newcount\my@repeat@count
\newcommand{\myrepeat}[2]{%
  \begingroup
  \my@repeat@count=\z@
  \@whilenum\my@repeat@count<#1\do{#2\advance\my@repeat@count\@ne}%
  \endgroup
}
\title{\LARGE \bf
Conditional GANs for Sonar Image Filtering with Applications to Underwater Occupancy Mapping
}

\author{Tianxiang Lin, Akshay Hinduja, Mohamad Qadri, and Michael Kaess % <-this % stops a space
\thanks{The authors are with The Robotics Institute, Carnegie Mellon University, USA. {\tt\small \{tianxian, ahinduja, mqadri, kaess\} @andrew.cmu.edu}}%
\thanks{This work was partially supported by the Office of Naval Research grant N00014-21-1-2482.}
\thanks{The authors would like to thank Easton Potokar for sharing real-world sonar data used for training our network.}
}

\begin{document}

\maketitle
\thispagestyle{empty}
\pagestyle{empty}

%%%%%%%%%%%%%%%%%%%%%%%%%%%%%%%%%%%%%%%%%%%%%%%%%%%%%%%%%%%%%%%%%%%%%%%%%%%%%%%%
\begin{abstract}

Underwater robots typically rely on acoustic sensors like sonar to perceive their surroundings. However, these sensors are often inundated with multiple sources and types of noise, which makes using raw data for any meaningful inference with features, objects, or boundary returns very difficult. While several conventional methods of dealing with noise exist, their success rates are unsatisfactory. This paper presents a novel application of conditional Generative Adversarial Networks (cGANs) to train a model to produce noise-free sonar images, outperforming several conventional filtering methods. Estimating free space is crucial for autonomous robots performing active exploration and mapping. Thus, we apply our approach to the task of underwater occupancy mapping and show superior free and occupied space inference when compared to conventional methods.    

\end{abstract}

%%%%%%%%%%%%%%%%%%%%%%%%%%%%%%%%%%%%%%%%%%%%%%%%%%%%%%%%%%%%%%%%%%%%%%%%%%%%%%%%
\section{Introduction}
Autonomous Underwater Vehicles (AUVs) are useful in a broad range of applications that are otherwise tedious or potentially dangerous for humans to perform. A major aspect of AUVs is to perform underwater mapping which assists in tasks like subsea infrastructure inspection, ship hull inspection \cite{Kaess2010TowardsAS}, seafloor surveying, and bathymetry\cite{fallon11icra}. 

In most conditions, AUVs cannot rely on optical sensors like cameras and laser range scanners due to limitations arising from turbidity and light absorption. Visibility in deep water is often constrained to a few meters (1-2 meters) at best. Thus, acoustic sensors like sonar are better suited for tasks involving mapping and free space estimation, often providing range information from several meters (10 meters and up) depending on the type and frequency of the sonar. 

Different tasks need different sonar types. For seafloor mapping, side scan sonar (SSS) is commonly used as in~\cite{chen11Oceans,fallon11icra}, whereas in more structured and complex 3D environments, imaging sonars are found to be more prevalent as shown in~\cite{Hernandez09ifac,Cho17joe,Marani09oe,Hover12ijrr}. Teixeira et al. \cite{Teixeira16iros} also used an imaging sonar with a concentrator lens in profiling mode, utilizing submaps to obtain occupancy and subsequently perform simultaneous localization and mapping (SLAM).

\begin{figure}[h!]
\centering
  \includegraphics[width=0.9\columnwidth]{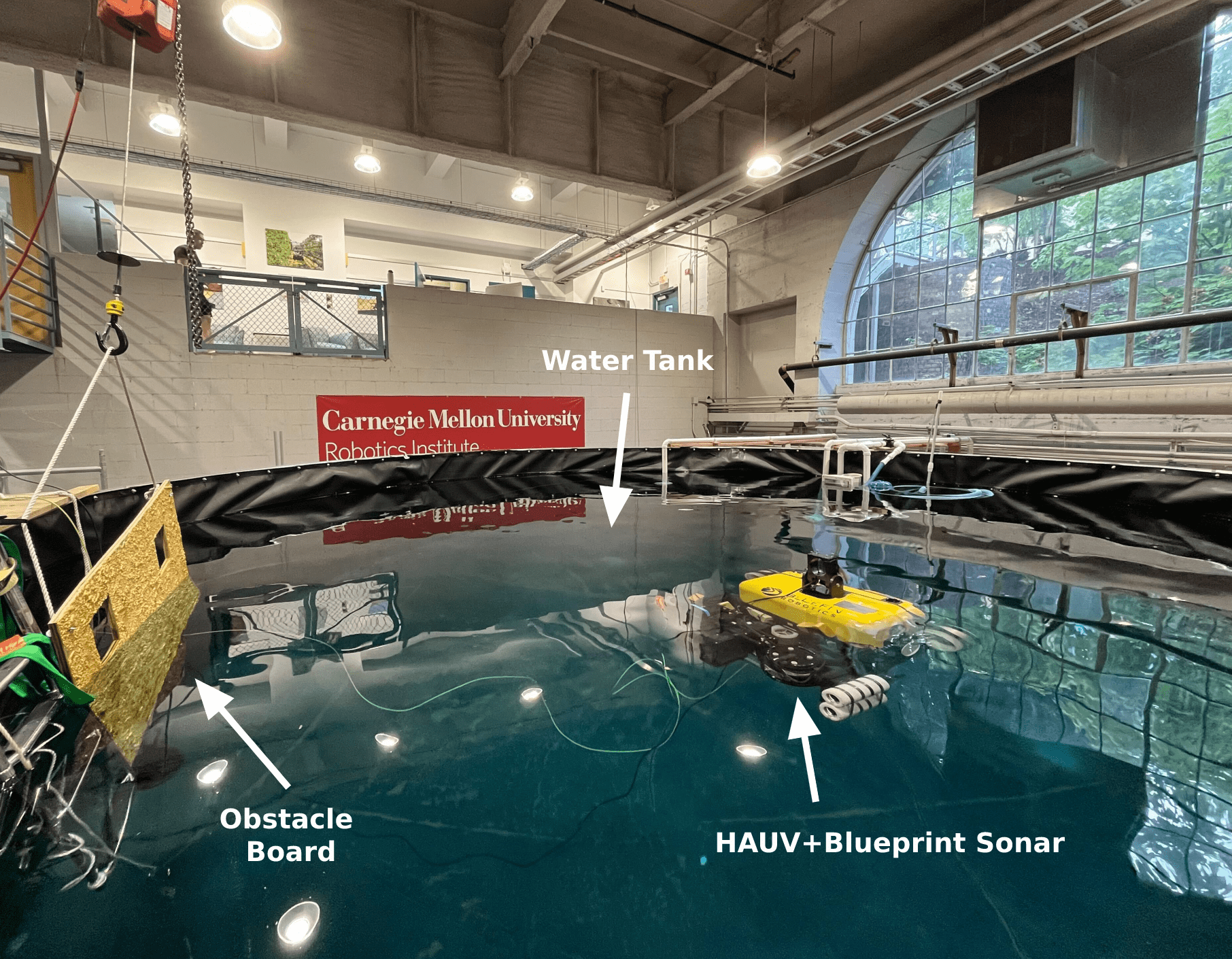}
  \caption{Real-world experiments performed in a test tank.}
  \label{tank-exp}
\end{figure}

\begin{figure}[h!]
\centering
  \includegraphics[width=1\columnwidth]{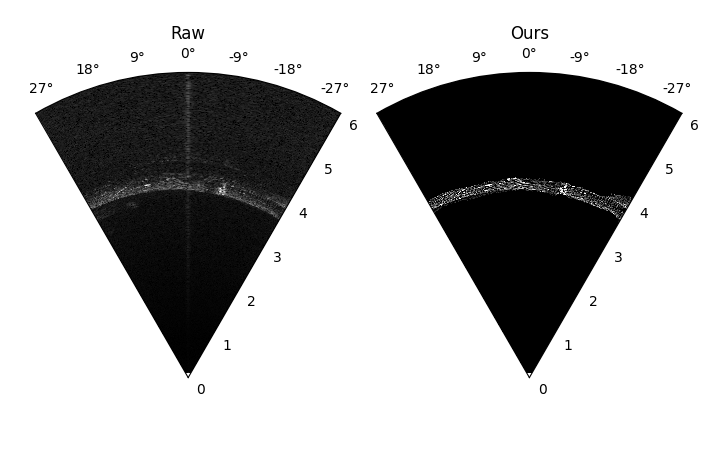}
  \caption{Real-world sonar image filtering by our method.}
  \label{filter_ex}\vspace{-4mm}
\end{figure}

While sonar is the preferred sensing modality for underwater environments, it is far from ideal due to noise from random and systematic errors. The sources for these errors can be due to the environment, temperature, an imperfection in calibration, cross-talk between transducers and receivers, multipath reflections, as well as delayed signal returns during vehicle motion~\cite{VanMiddlesworthThesis}. The most common noise pattern observed is speckle noise. It mainly originates from coherent and random patterns of constructive and destructive interference of backscattered signals creating pixels with low and high intensity in the image. This is mainly due to wave propagation characteristics and the material properties of the objects in the frustum of the sonar. It is also a primary source of visual noise for many wave propagation-based sensors~\cite{Jaybhay2015sip,Karabchevsky2011Oceans,grabek2019Sensors}.

Reduced noise in sonar images is important for AUVs. It enables the robot to infer its environment better, through observing features or objects and performing SLAM. It is also important in the estimation of free space, as noise in the image can be incorrectly defined as an occupied space. Conventional speckle noise filters operate on the premise that the actual signal and noise are statistically independent and can be differentiated on this basis. While most conventional filters are capable of performing some degree of noise reduction, often it is at the cost of degradation of important image features such as corners and high contrast boundaries. % and their accuracy is limited. 

The remainder of the paper is organized as follows. After discussing related work in the next section, Section~\ref{statement} introduces imaging sonar geometry in both Cartesian and polar space. Section~\ref{method} subsequently describes our filtering method, followed by an evaluation of the filter performance in Section~\ref{Sec:Filter_results}. In Section~\ref{Occupancy results} we apply our method toward underwater occupancy mapping and show that our method outperforms conventional methods used for sonar denoising for both simulated and real-world sonar image data. Finally, Section~\ref{conclusions} closes with our concluding remarks and future directions for this work. 

\section{Related Work}
There have been several attempts at filtering sonar images through a multitude of methods. Some of the earliest attempts have been the Lee, Kuan, and Frost filters for synthetic aperture sonar images as described by \cite{Mascarenhas97}. The filters' measure of image homogeneity are based on the multiplicative speckle model, which is not always accurate causing failures for small details in images like corners. Lopes et al.~\cite{Lopes1990Enhanced} improved upon these methods by adding two thresholds on the coefficient of variation, allowing better homogeneous averaging and heterogeneous feature preservation. Variations of wavelet transforms have also performed well in denoising sonar images as seen in~\cite{isar2009ivp,sendur2002tsp}. The transforms work by concentrating the original signal and image features in a few large-magnitude wavelet coefficients, where the smaller magnitude coefficients represent the noise. Negahdripour et al.~\cite{Negahdaripour05crv} proposed using multiple images taken from a single viewpoint and averaging to obtain refined intensity returns. Anisotropic diffusion has shown success for feature-based methods~\cite{Shin15Oceans,Westman18icra} as well as for dense 3D reconstruction using sonar~\cite{Westman19iros}. Teixeira et al.~\cite{Teixeira16iros} utilize a two-dimensional Wiener filter~\cite{Hundt1980tsu} to deconvolve a custom point spread function for their sonar. Despite the variety of filtering algorithms, there is no consensus on what method is best since performance varies with the application of interest.

More recently, machine learning-based methods have gained popularity for general image denoising. Lu et al.~\cite{lu2019icmlc} use a deep convolutional network to reduce speckle noise. Their results show a great improvement for SSS images. Imaging sonars, especially forward-looking sonars would typically only have partial occupation from the viewed object with the rest of the image being empty with noise. Whether this network could handle this phenomenon is unknown.

The work of Isola et al.~\cite{pix2pix2017}, pix2pix, showed how a conditional adversarial network can be used for image-to-image translation, performing tasks like generating photos from edge maps, style transfer, and background removal to name a few examples. For applications to underwater imaging sonar, Lee et al.~\cite{Lee22ploso} have recently used pix2pix to produce imitation sonar images to augment their training data for a fully convolutional object segmentation network.

Our proposed method utilizes pix2pix in a different way. We train the conditional Generative Adversarial Network (cGAN) to generate binary masks of the actual returns. These masks when used to segment the original raw image help preserve the high-intensity objects detected in our sonar image, while eliminating most of the noise in the process as seen in Fig.~\ref{filter_ex}. The effectiveness of our method is showcased by using the filtered images to construct an occupancy map.

\section{Background: Imaging Sonar Geometry}\label{statement}

\begin{figure}[hb!]
\centering
  \includegraphics[width=\columnwidth]{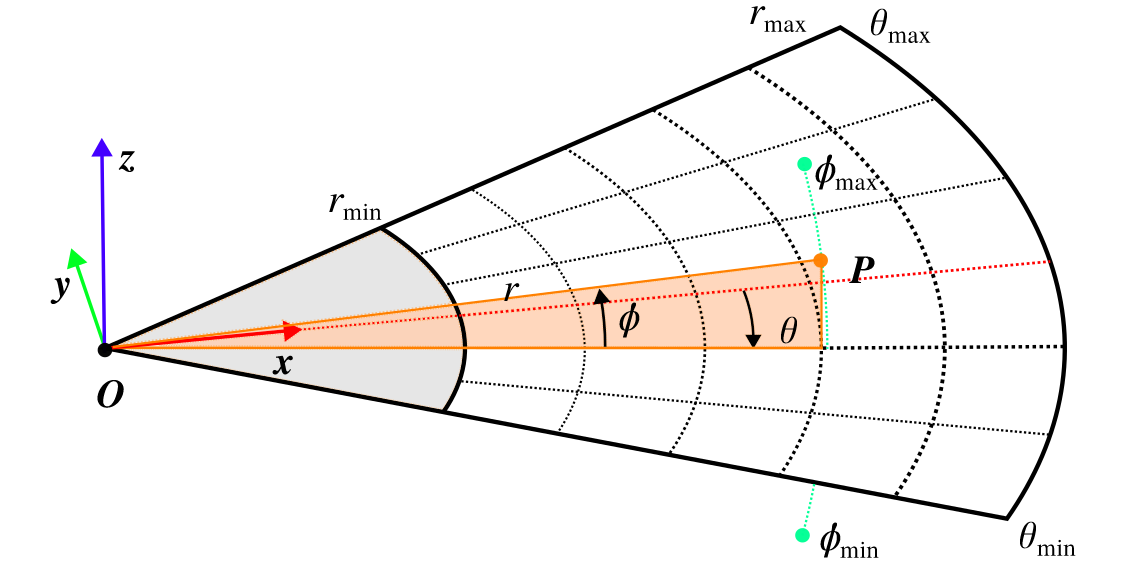}
%   \caption{Geometry of a single sonar image. Point $\mathbf{P}$ is represented by range $r$, elevation $\phi$, and bearing $\theta$. $r_{\text{max}}$ and $r_{\text{min}}$ are the maximum and minimum ranges from the imaging sonar. $\phi_{\text{max}}$ and $\phi_{\text{min}}$ are the maximum and minimum elevation angles. $\theta_{\text{max}}$ and $\theta_{\text{min}}$ are the maximum and minimum azimuth angles from the imaging sonar.}
  \caption{Geometry of a single sonar image. Point $\mathbf{P}$ is represented by range $r$, elevation $\phi$, and bearing $\theta$. $r_{\text{max}}$, $r_{\text{min}}$, $\phi_{\text{max}}$, $\phi_{\text{min}}$, $\theta_{\text{max}}$, and $\theta_{\text{min}}$ are respectively the maximum and minimum ranges, elevation angles, and azimuth angles of the imaging sonar.}
  \label{sonar_geometry}
\end{figure}

Consider a point $\boldsymbol{P}(\theta, r, \phi)$ in the field of view of the imaging sonar, parameterized in the local spherical sonar coordinate system as seen in Fig.~\ref{sonar_geometry}. Here, $\theta$, $r$, and $\phi$ represent the bearing, range, and elevation angle of the point respectively. The conversion of $\boldsymbol{P}$ to the Cartesian frame to point $\boldsymbol{C}(x, y, z)$ and vice versa is
\begin{gather}
    \boldsymbol{C} = \begin{bmatrix}
        x\\ y\\ z
    \end{bmatrix} =
    \begin{bmatrix}
        r \cos \theta \cos \phi\\
        r \cos \theta \sin \phi\\
        r \sin \theta
    \end{bmatrix}\label{eq:xyz}
\end{gather}

\begin{gather}
    \boldsymbol{P} = \begin{bmatrix}
        \theta\\ r\\ \phi
    \end{bmatrix} =
    \begin{bmatrix}
        \arctan2(y,x)\\
        \sqrt{x^2+y^2+z^2}\\
        \arctan2(z,x^2+y^2)
    \end{bmatrix}\label{eq:phirtheta}
\end{gather}

Imaging sonars generate partial spherical measurements by sending out acoustic signals into a frustum. Time of flight measured from the reflected signals observed by the transceivers provide the range $r$ and bearing $\theta$, of the reflecting surface. However, these measurements are unable to disambiguate the elevation $\phi$ of the reflected signal. Due to this, all detected returns from a single elevation arc project onto the same pixel of a range image $\boldsymbol{I}(\theta, r)$. For a pixel corresponding to a certain bearing and range in $\boldsymbol{I}$, the pixel's intensity corresponds to the intensity of the reflected signal from all returns along the elevation.  

\begin{figure}[h!]
\centering
  \includegraphics[width=\columnwidth]{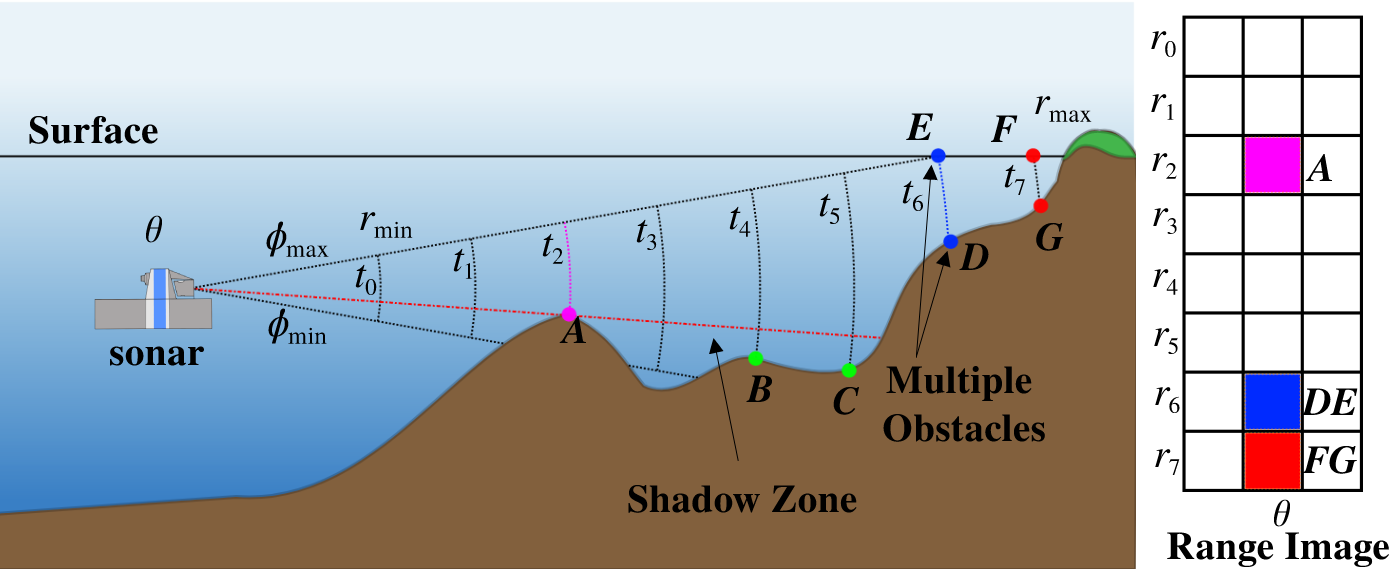}
  \caption{Elevation ambiguity and shadow zone of imaging sonar. Green points, B and C, intercepted by sonar lie in the shadow zone and are thus not visible in the generated range image. Blue points (D and E) and red points (F and G), lying on the same elevation arc, cannot be distinguished in the image.}
  \label{elevation_ambiguity}
\end{figure}

Along with elevation ambiguity, shadow zones also appear when objects closer to the sonar obstruct the view of obstacles behind them. Hence, low pixel intensity values in the sonar images do not necessarily mean the absence of obstacles. A pictorial representation of shadow zones and elevation ambiguity can be seen in Fig.~\ref{elevation_ambiguity}

\section{Sonar Image Filtering with cGANs}\label{method}

We present an approach which uses cGANs to filter imaging sonar images. GANs consist of two networks, a generator $G$ and a discriminator $D$. The generator's task is to produce data that closely matches the training distribution and the discriminator has the task of distinguishing real images from those produced by the generator. The generator in GANs tries to learn the underlying training image distribution with no additional constraints. Hence, the generators are free to learn any mapping from the noise distribution $p(z)$ to $p_{data}$, which are the sonar images and corresponding mask pairs. 

Pix2pix, like other cGANs, conditions the output of the generator model on an input image through the discriminator model. In cGANs, the training data is in the form of a pair $(x,y)$, where $x$ and $y$ represent the original image and labeled image that $x$ is conditioned on respectively. The training objective can be represented as
\begin{gather}
    {Loss} = \mathop{\text{min}}_{G}\mathop{\text{max}}_{D}\mathcal{L}_{c}(G,D) + \lambda\mathcal{L}_{L1}(G)
        \label{eq:pix2pix_loss}
\end{gather}
where $\lambda$ is the weight applied to the L1 loss term, and the objective of cGANs $\mathcal{L}_{c}$ and L1 loss term $\mathcal{L}_{L1}$ are
\begin{multline}
    \mathcal{L}_{c}(G,D) = E_{x,y\sim p_{data}}[\text{log}~D(x,y)] \\
    +E_{x\sim p_{data},z \sim p(z)}[\text{log}~(1-D(x,G(x,z)))]
        \label{eq:Lc_loss}
\end{multline}
\begin{gather}
    \mathcal{L}_{L1} = E_{x,y\sim p_{data},z\sim p(z)}\lVert {y-G(x,y)} \rVert_{L1}
        \label{eq:L1_loss}
\end{gather}

The discriminator's task in cGANs remains the same, which is to differentiate real image pairs $(x,y)$ from fake pairs $(x,G(x,z))$. The added task for the generator is to make sure the produced images are nearer to the label $y$, encouraged by the loss term $\mathcal{L}_{L1}$.

Our proposed work provides noisy sonar images (as $x$) and binary masks (as $y$) to pix2pix to generate close-to-ground truth binary masks as an output. This approach is in inverse to Lee et al.~\cite{Lee22ploso}, who give binary masks (as $x$) and noisy sonar images (as $y$) to the generator to get realistic imitations of sonar images.

\begin{figure}[h]
\centering
  \includegraphics[width=\columnwidth]{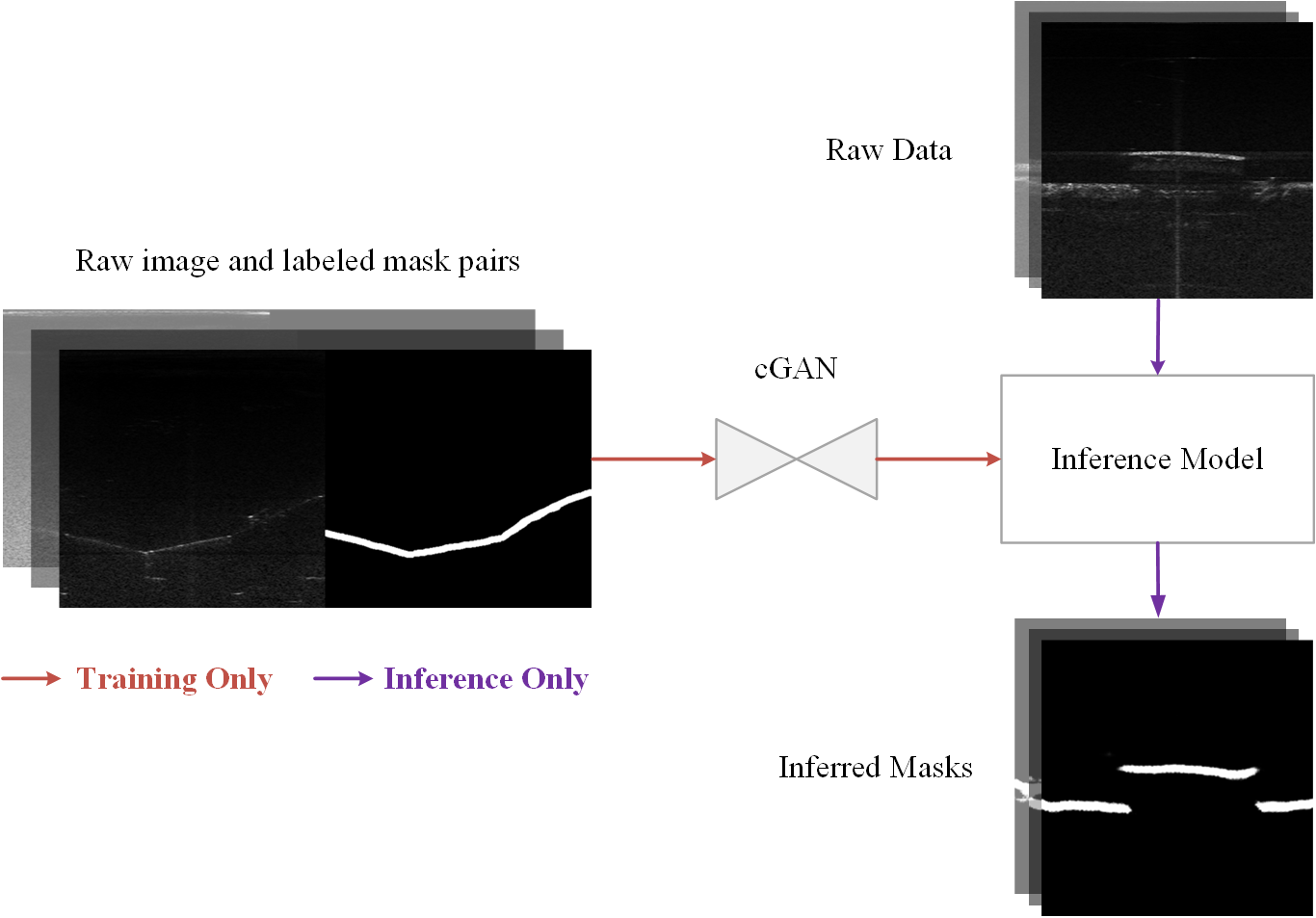}
  \caption{Conditional GAN for filtering sonar images. Raw and labeled mask pairs are used to train the network. Passing raw data through the network gives a binary mask of the segmented returns. }
  \label{pix2pix}
\end{figure}

Fig.~\ref{pix2pix} describes our training procedure of how the cGAN is trained to obtain the inferred mask to filter the raw image for real-world data. We provide a series of noisy raw data as the input image, $x$, and a binary mask of the actual surfaces for the same scenes as our target image, $y$. When $G$ and $D$ are trained together with a sufficient number of image pairs, we get a network that produces a well-defined binary image mask as our output for raw input data. This mask is then used to segment out the actual return in the sonar image while eliminating a majority of the image noise. As the generated mask typically has a buffer zone enveloping the object in view, an unsharp filter is then applied to the trained model's outputs to further contrast the reflected signal from the surrounding residual noise.  

\section{Filtering Results}\label{Sec:Filter_results}

In this section, we describe our prepared dataset and training parameters before presenting results for both simulation and real-world sonar images.
For our simulated data, we use the HoloOcean simulator~\cite{potokar22icra,Potokar22iros} and for our real-world data, we use a Bluefin Hovering Autonomous Underwater Vehicle (HAUV)~\cite{BluefinHAUV} equipped with a 1.2MHz Teledyne/RDI Workhorse Navigator Doppler velocity log (DVL), a Honeywell HG1700 inertial measurement unit (IMU), and a BluePrint Subsea M1200d imaging sonar~\cite{Blueprint} as seen in Fig.~\ref{tank-exp}. The DVL/IMU navigation solution has a time-based drift, estimated to be 1.1 meters in 20 minutes. We train two models, one for simulated data, and the other for real-world data separately. 

To train our simulated data model, we generate 3000 frame pairs of noisy raw data, and noise-free images converted to binary masks. For the real-world data, we randomly select 100 frames from a data log, and manually prepare binary mask images to serve as the target frames. While the HoloOcean simulator is modeled on the M1200d, we observed different noise patterns in simulation and real data, in part due to the small, enclosed nature of the test tank which amplified multipath reflections. Thus attempts to fine-tune a network trained on simulated data with real-world data were unsuccessful. However, a separately trained model worked exceedingly well, even with a modest amount of hand-labeled image pairs. Images used for training were from different environments than those used for testing, for both simulation and real-world data. 

Both models were trained using pix2pix~\cite{pix2pix2017} with an Intel(R) Core(TM) i7-7820X CPU running at 3.60GHz and a NVIDIA GeForce RTX 3080 Ti GPU for training. The training time observed for 3000 simulated image frames was roughly 4.5 hours. 
% python 3.8.13
% torch 1.9.0+cu111
% Intel(R) Core(TM) i7-7820X CPU @ 3.60GHz
% NVIDIA GeForce RTX 3080 Ti 12 GB
% 4.5 h training

\begin{figure}[ht!]
\centering
  \includegraphics[width=\columnwidth]{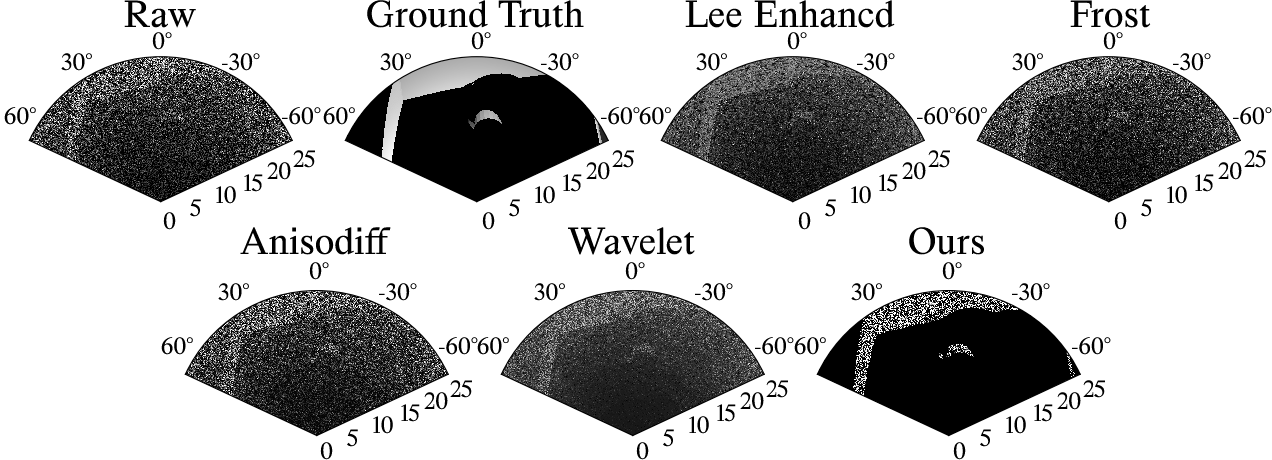}
  \caption{Filtering results on simulated sonar data with added speckle noise. The enhanced Lee and wavelet filter perform best for conventional filters. Our method is able to remove background noise efficiently and give well-defined borders for the returned signal. }
  \label{range_images_holo}
\end{figure}

\begin{figure}[hbt]
\centering
  \includegraphics[trim={0 0 0 1cm},clip, width=\columnwidth]{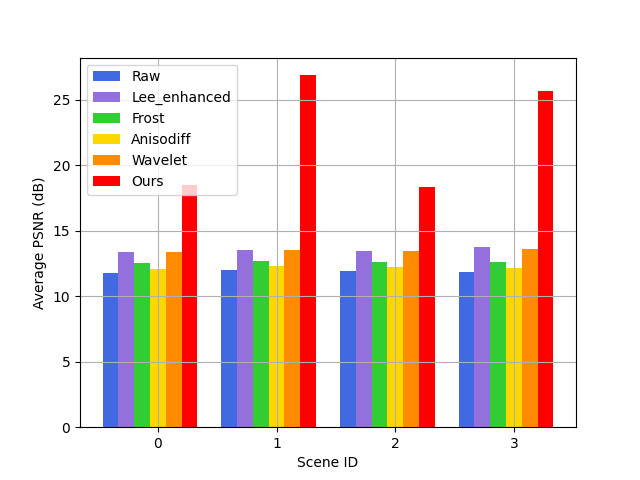}
  \caption{Comparison of PSNR of our and selected conventional methods on four different scenes from the HoloOcean simulator. Our method consistently shows higher PSNR values for all scenes.}
  \label{psnr_pierharbor}
\end{figure}

\subsection{Simulation Results}
% Filters Parameters:
% Lee enhanced: window_size=5, k_value=1.0 (default), cu_factor=0.25 (default), cmax=1.73 (default)
% Frost: window_size=5, damping_factor=1.0
% Anisodiff: number of iterations=2, kappa (conduction coefficient)=60, gamma (max value of .25 for stability)=0.1, step (the distance between adjacent pixels in (y,x)) = (1.,2.),  option= 1 (Perona Malik diffusion equation No 1)
% Wavelet: soft threshold, sigma=estimated_sigma / 4 (estimated_sigma: Robust wavelet-based estimation of the (Gaussian) noise standard deviation.)

We compare our method against four filters: 1. the original Frost filter, 2. the enhanced Lee filter, 3. Anisotropic diffusion, and 4. wavelet transforms using the VisuShrink threshold~\cite{Donoho94idealspatial}. Parameters were adjusted to attain the best performance of each filter.

% (TODO: Tianxiang mentions parameters)
We choose the peak signal-noise ratio (PSNR) as the error metric to compare the filtration results. As speckle noise is generally formed due to the constructive and destructive interference of reflected signals from largely uneven surfaces, it has been observed that HoloOcean's simulated images were unable to replicate the effect due to the smooth simulated surfaces on the environment models. To make our comparison more realistic we add normally distributed speckle noise to all images across four scenes from HoloOcean's list of available environments. Fig.~\ref{range_images_holo} shows the qualitative comparative results of filtration of a sample frame. We note that our method provides a much sharper distinction of object boundaries, while conventional filters cannot achieve so to a satisfactory degree. In Fig.~\ref{psnr_pierharbor} we compare the baseline filters and our method through the average PSNR per scene. Across all the scenes, our method produces a higher PSNR value, indicating better performance and robustness to noise. 

\begin{table}[hbt]
\centering
\caption{Average run time per frame for filtering methods.}
\label{tab:time}
\resizebox{\columnwidth}{!}{%
\begin{tabular}{|c|ccccc|}
% \hline
\hline
\textbf{Filters}            & \textbf{Lee Enhanced} & \textbf{Frost} & \textbf{Anisodiff} & \textbf{Wavelet} & \textbf{Ours} \\ \hline
\textbf{Time per frame (s)} & 8.75                  & 22.54          & 0.03               & 0.04             & 0.56          \\ \hline
\end{tabular}%
}
\end{table}

We also compare the time taken for obtaining a filtered image from raw data for different methods as seen in Table.~\ref{tab:time}. All methods were run on the CPU. While our method takes significantly more time for inference compared to methods such as the wavelet transform and anisotropic diffusion, the higher filtering performance is more desirable. When it comes to slow-moving Autonomous Underwater Vehicles (0.5-1.5 knots~\cite{BluefinHAUV}), the time taken per frame is not as detrimental. 

\subsection{Real World Results}

\begin{figure}[hbt!]
\centering
  \includegraphics[width=\columnwidth]{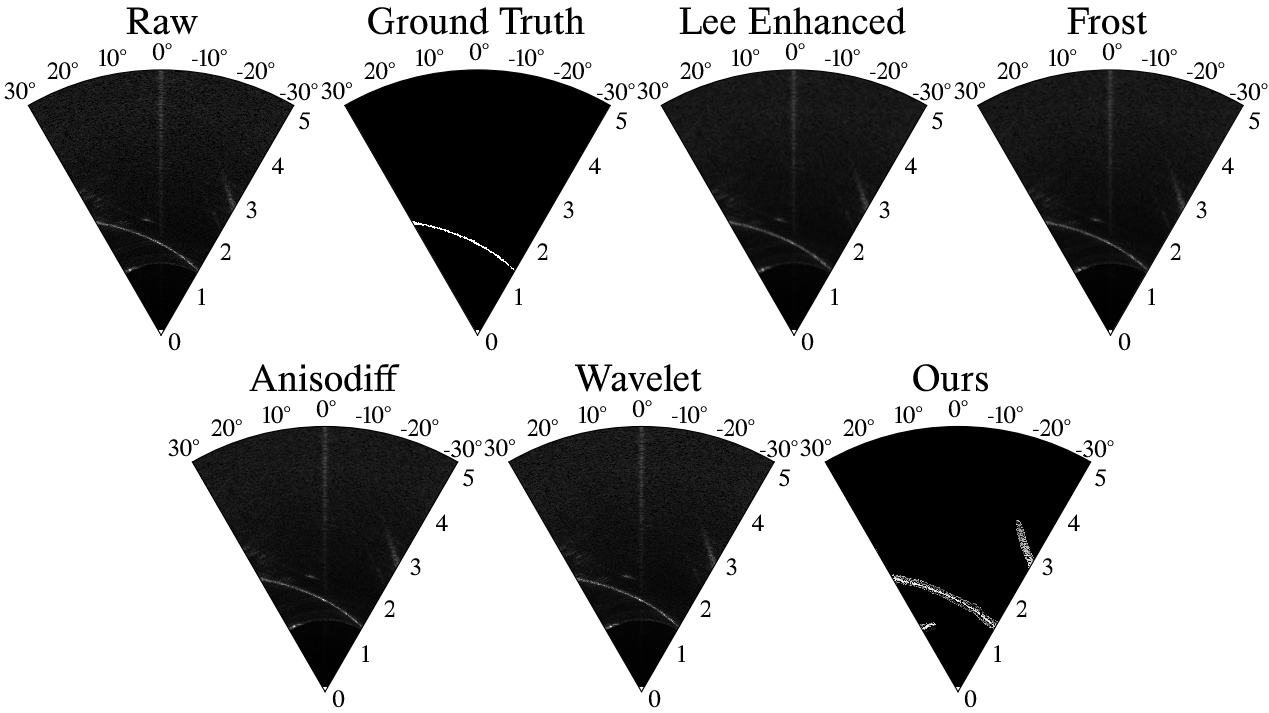}
  \caption{Filtering results on real-world sonar images from test tank experiments. Ground truth image is prepared by manually removing noise and sharpening the actual return from the tank wall. While contours resulting from multipath propagation exist, our result filters out most noise compared to other conventional methods.}
  \label{real_sonar}
\end{figure}

For our real-world data, we use the Blueprint Oculus M1200d in its high frequency (2.1MHz) mode which gives a minimum range of 0.1 meters and is configured to a max range of 5 meters, with a $60\degree$ horizontal field of view. Fig.~\ref{real_sonar} compares the filtration performance of our model for a single frame versus the same aforementioned conventional filters. We see that our method is better able to filter surrounding noise and give a sharp contrast to the returned signal. 

% Ray-casting with range images
\begin{figure*}[t]
\centering
  \includegraphics[width=\textwidth]{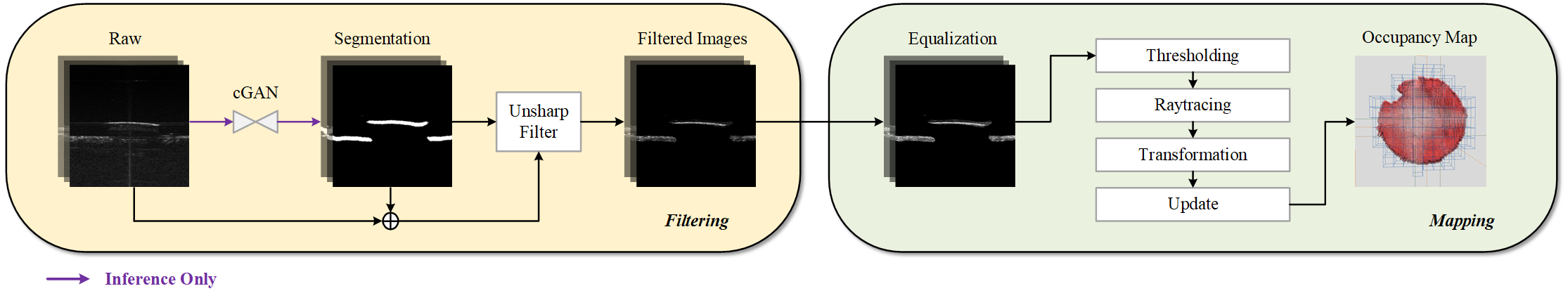}
  \caption{Underwater occupancy mapping framework. Raw images passed through the network produce a binary mask which is used to segment the original raw image. After passing through an unsharp filter, the image goes through a histogram equalization to generate a higher contrast image for thresholding. Using vehicle poses, we obtain our final occupancy map.}
  \label{mapping}
\end{figure*}

\section{Application to Occupancy Mapping and Evaluation} \label{Occupancy results}

Before an AUV can infer its surroundings for mapping purposes, the estimation of free space is important. For active exploration and mapping, a robot must know the obstacle-free space surrounding itself to plan safe and efficient paths to
unexplored regions. Thus, in this section, we apply our method to underwater occupancy mapping. 
The complete framework we use for underwater occupancy mapping with imaging sonar is shown in Fig.~\ref{mapping}. Raw sonar image frames are passed through the filtering module to provide denoised images which are then passed to the mapping module. With the filtered range images and known robot poses from the DVL/IMU, the Cartesian coordinates of obstacles can be computed using Equation~\ref{eq:xyz} and utilized in the 3D Bresenham line drawing algorithm~\cite{Liu2002} to update the values within the corresponding occupancy grid cells.

We begin with the inverse sensor model for our occupancy mapping framework, and then present and evaluate the occupancy maps generated with both simulation and real-world data. 

\subsection{Inverse Sensor Model}
% Grid Definition & % Inverse Sensor Model

Once we have the denoised sonar image, histogram equalization is performed~\cite{Blondel_2010} which increases the contrast, giving the regions of interest a higher intensity.
The intensity of corresponding pixels within sonar range images will determine the occupancy of the grid at the pixel location. We define the occupancy of a grid cell in an occupancy grid map $\mathbf{m}$ in the log-odd form as: 
\begin{align}
    {l}_{x,y} = \log \frac{p(\mathbf{m}_{x,y}|\mathbf{z})}{1-p(\mathbf{m}_{x,y}|\mathbf{z})}
    \label{logodd}
\end{align}
where $\mathbf{m_{x,y}}$ denotes a binary status of a grid cell at position $x,y$, and $\mathbf{z}$ are the measurements from the sensor from time $1$ through time $T$ \cite{thrun2005}. 

For the given sonar range image data, we place higher confidence on the estimated free grid cells rather than occupied cells. This is due to the elevation ambiguity observed in imaging sonars, and the possibility of cells estimated as occupied might not have a significant obstacle along the entire elevation arc. We model our occupancy using the following inverse sensor model:
\begin{equation}
  \textbf{inverse\_sensor\_model}(\mathbf{m}, \mathbf{x}, \mathbf{z}) = 
   \begin{cases}
        l_{\text{free}} & \text{if $\mathbf{z}$ $<$ $t$,}\\
        l_{\text{occ}} & \text{otherwise,}
    \end{cases} 
    \label{inverse_sensor_model}
\end{equation} where $\mathbf{m}$ represents the status of an occupancy grid cell, $\mathbf{x}$ denotes the state of the sensor itself, $\mathbf{z}$ is the measurement of intensity from sonar images, and $t$ is the threshold used to detect the obstacles within the range image. We use the following probability ratios for free and occupied cells, which performed the best for simulations: $p_{\text{free}} = 0.55$, and $p_{\text{occ}} = 0.05$ \cite{thrun2005}.

\begin{figure}[H]%
    \centering
    \begin{subfigure}{\columnwidth}
        \centering
        \includegraphics[width=0.9\columnwidth]{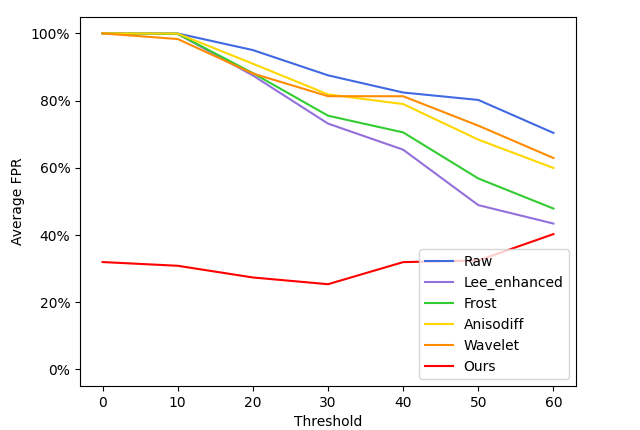}%
        \caption{Average False Positive Rate.}%
        \label{fig_fprs}%
    \end{subfigure}\hfill%
    \begin{subfigure}{\columnwidth}
        \centering
        \includegraphics[width=0.9\columnwidth]{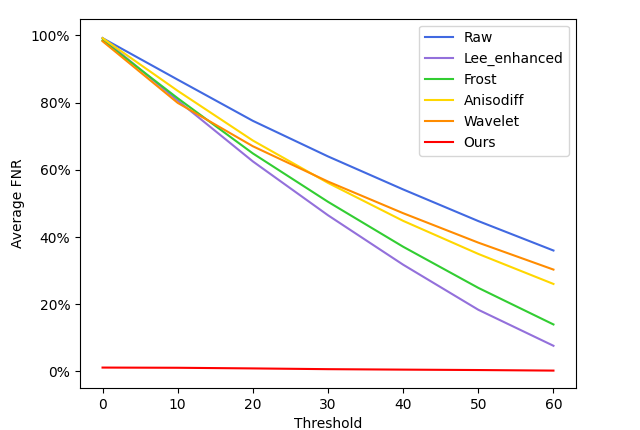}%
        \caption{Average False Negative Rate.}%
        \label{fig_fnrs}%
    \end{subfigure}\hfill%
    \caption{Average False Positive (a) and False Negative (b) rates for generated occupancy map using different filtering algorithms and pixel intensity thresholds for simulated data. }
    \label{fig_omap_eval}
\end{figure}

\subsection{Occupancy Map Evaluation for Simulation Data}

For measuring the accuracy of our occupancy map, we use the assumption that a positive cell value from our inverse model indicates that the grid cell is free. Hence, a false positive occurs when a grid cell is indicated as free but is actually occupied. Similarly, a false negative means that a cell is occupied while it is literally free. We do not make assumptions about unknown grid cells. Using this model we plot the average False Positive Rate (FPR) and False Negative Rate (FNR) for all scenes across different pixel intensity thresholds, $t$ from 0 to 60 for \texttt{uint8} images. These results are as seen in Fig.~\ref{fig_omap_eval}. As anticipated, our method performs the best across all threshold values. As the threshold increases, the enhanced Lee filter gets closer to the performance of our method. But given its filtration run-time, it is not good for online estimation of free space.

\begin{figure}[h!]%
    \centering
    \begin{subfigure}{.45\columnwidth}
        \includegraphics[width=\columnwidth]{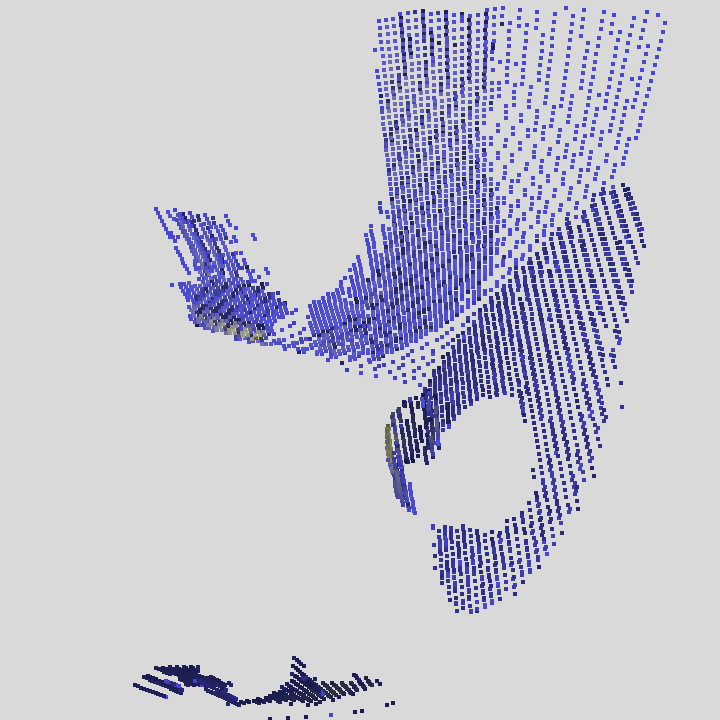}%
        \caption{Ground Truth}%
        \label{holo_gt}%
    \end{subfigure}\hfill%
    \begin{subfigure}{.45\columnwidth}
        \includegraphics[width=\columnwidth]{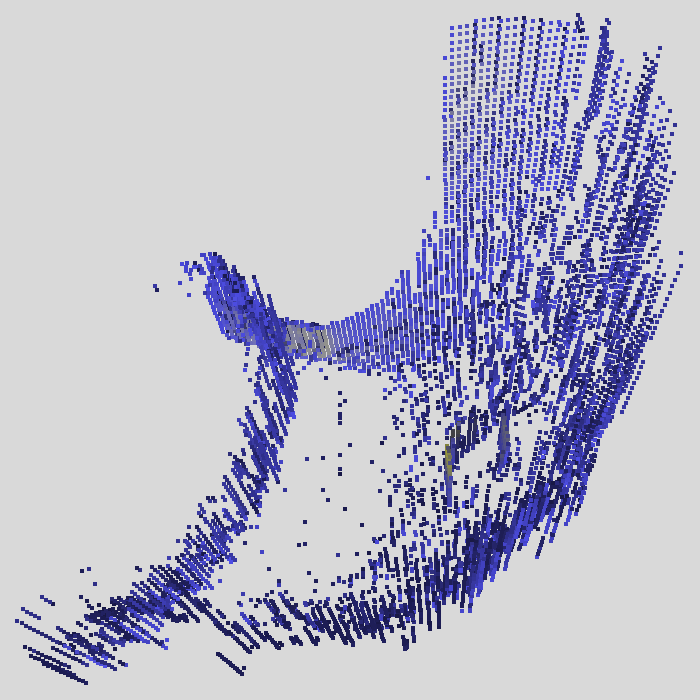}%
        \caption{Raw}%
        \label{holo_raw}%
    \end{subfigure}\hfill%
    \begin{subfigure}{.45\columnwidth}
        \includegraphics[width=\columnwidth]{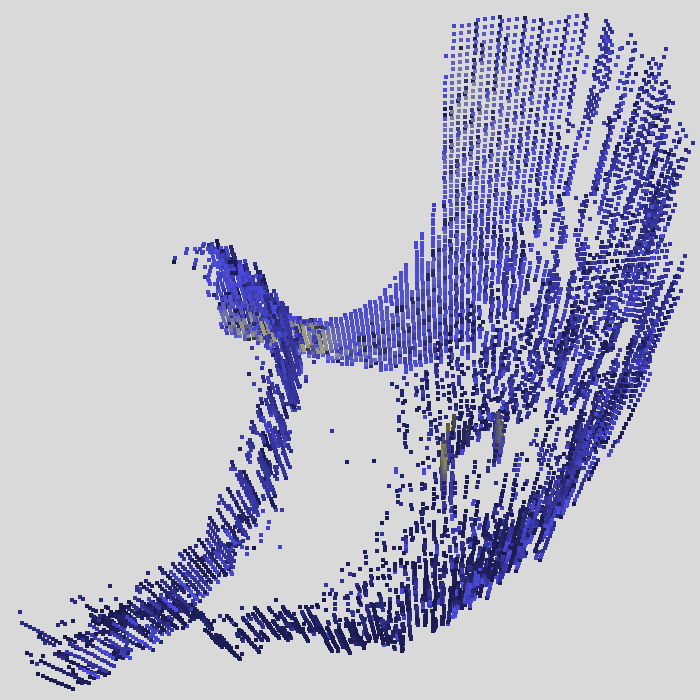}%
        \caption{Wavelet}%
        \label{holo_wavelet}%
    \end{subfigure}\hfill%
    \begin{subfigure}{.45\columnwidth}
        \includegraphics[width=\columnwidth]{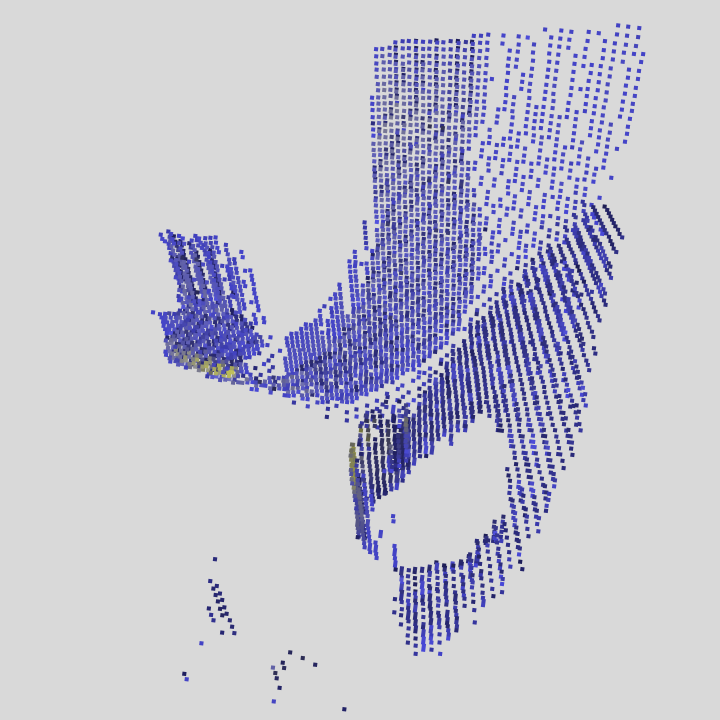}%
        \caption{Ours}%
        \label{holo_pix2pix}%
    \end{subfigure}\hfill%
    \caption{Final occupancy maps obtained from different filtering strategies using simulated data}
    \label{omap_holo}
\end{figure}

The qualitative results of the generated occupancy map for the wavelet transform and our method compared to the ground truth and raw map are shown in Fig.~\ref{omap_holo}, where we see that our method can provide better free and occupied space information compared to the wavelet transform. We demonstrate the result from the wavelet transform since the chosen method performs the best among conventional filtration approaches for simulations.

\subsection{Occupancy Map Evaluation for Real-world Data}

\begin{figure}[hbt!]%
    \centering
    \begin{subfigure}{.49\columnwidth}
        \includegraphics[width=\columnwidth]{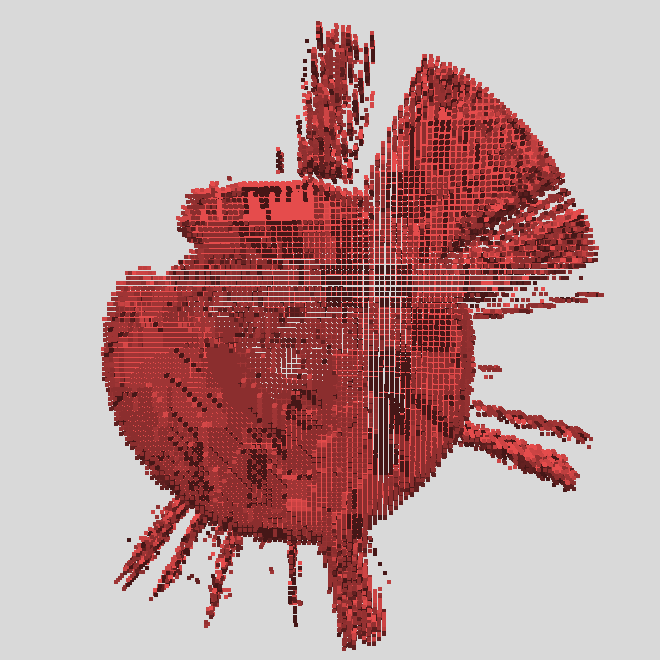}%
        \caption{Wavelet transform free space}%
        \label{real_wavelet_free}%
    \end{subfigure}\hfill%
    \begin{subfigure}{.49\columnwidth}
        \includegraphics[width=\columnwidth]{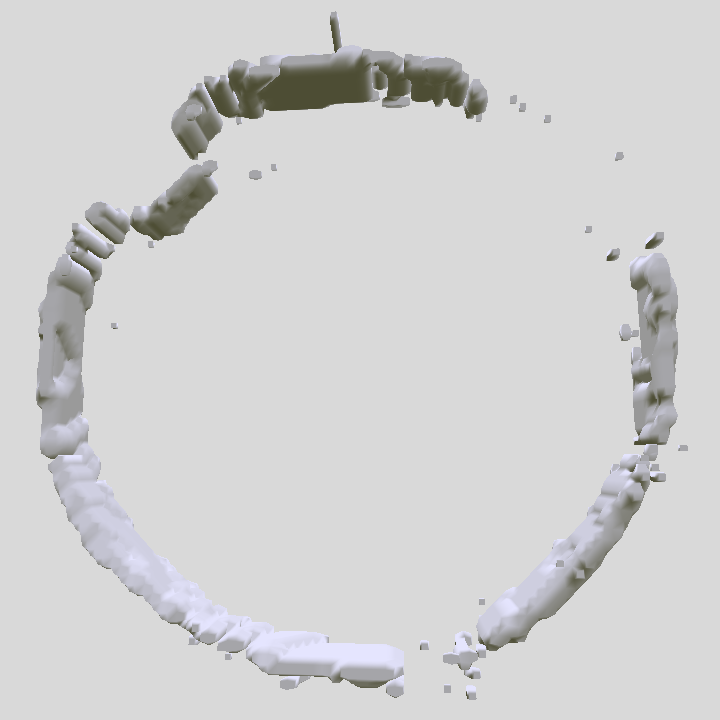}%
        \caption{Wavelet transform occupied space}%
        \label{real_wavelet_occ}%
    \end{subfigure}\hfill%
    \begin{subfigure}{.49\columnwidth}
        \includegraphics[width=\columnwidth]{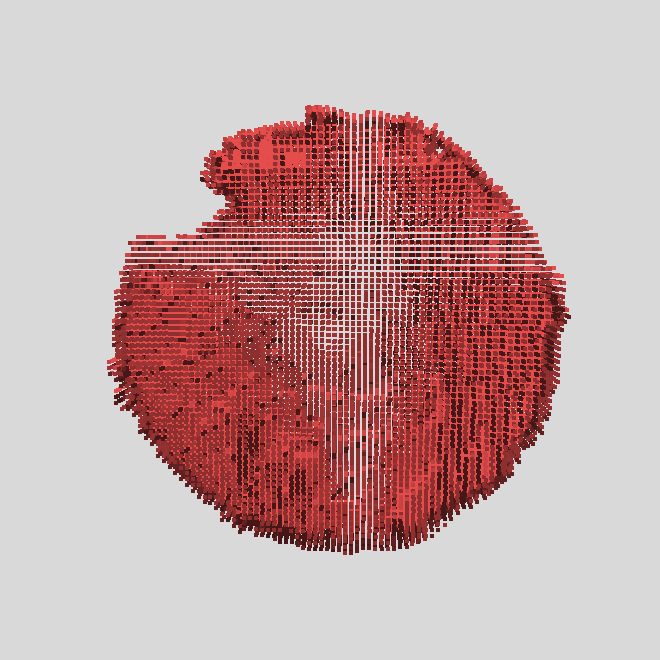}%
        \caption{Ours, free space}%
        \label{real_raw_free}%
    \end{subfigure}\hfill%
    \begin{subfigure}{.49\columnwidth}
        \includegraphics[width=\columnwidth]{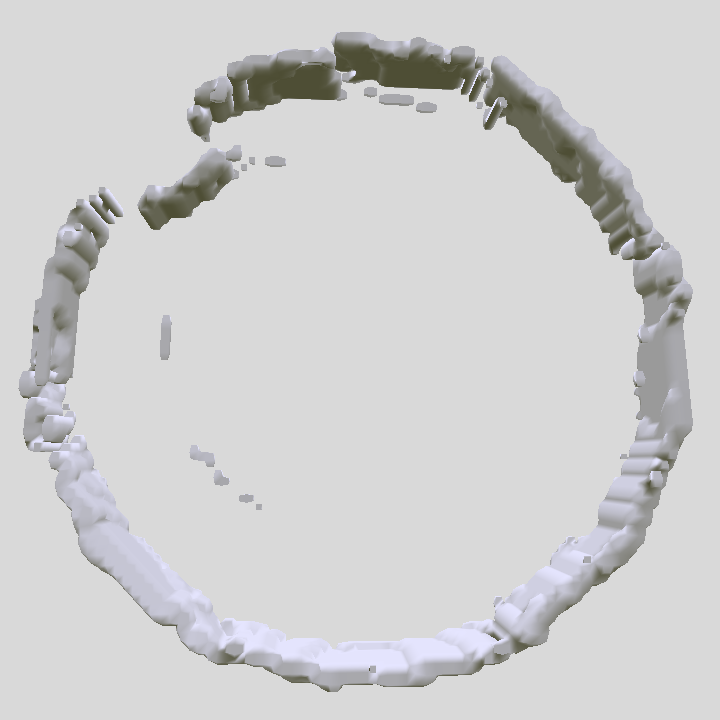}%
        \caption{Ours, occupied space}%
        \label{real_raw_occ}%
    \end{subfigure}\hfill%
    \caption{Occupancy maps generated through wavelet filtering and our method from real data in the test tank. Our method is able to infer available free space more accurately}
    \label{omap_real}
\end{figure}

Real-world data collected from our test tank, where a static obstacle board was placed adjacent to the tank wall, as seen in Fig.~\ref{tank-exp} was used to qualitatively compare the free and occupied space maps generated by our method and the wavelet transform baseline as seen in Fig.~\ref{omap_real}. We find that our method is more accurate for the experiment environment in determining both, free and occupied cells.

\begin{figure}[hbt!]%
    \centering
    \begin{subfigure}{.33\columnwidth}
        \includegraphics[width=\columnwidth]{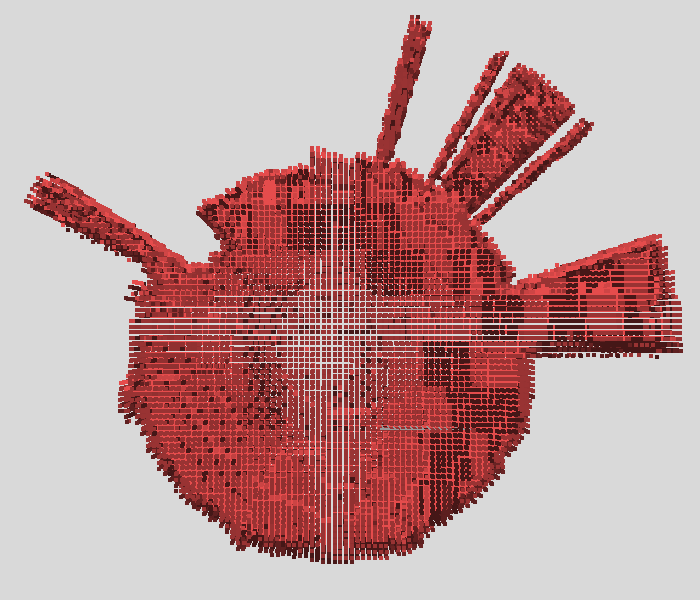}%
        \caption{25 pairs}%
        \label{fig_thresh_25}%
    \end{subfigure}\hfill%
    \begin{subfigure}{.33\columnwidth}
        \includegraphics[width=\columnwidth]{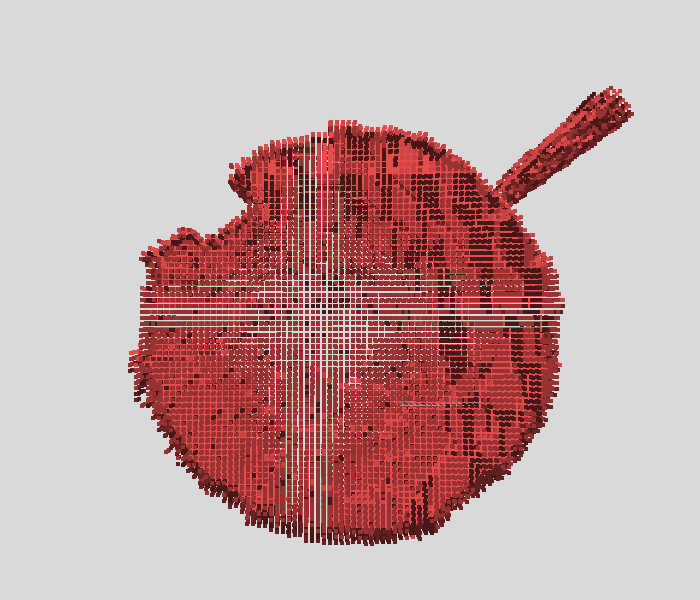}%
        \caption{50 pairs}%
        \label{fig_thresh_50}%
    \end{subfigure}\hfill%
    \begin{subfigure}{.33\columnwidth}
        \includegraphics[width=\columnwidth]{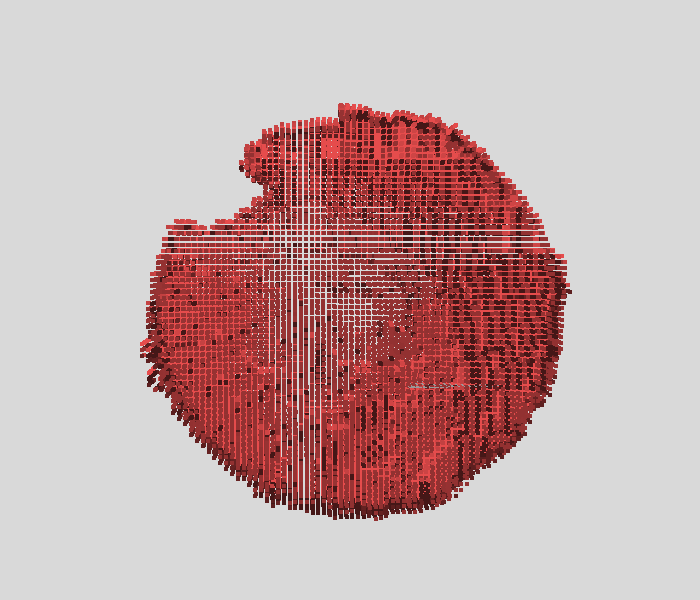}%
        \caption{100 pairs}%
        \label{fig_thresh_100}%
    \end{subfigure}\hfill%
    \caption{Occupancy maps by our filter using Conditional GAN. Each model is trained by different pairs of raw data and hand-labeled masks.}
    \label{fig_omap_real_models}
\end{figure}

We also experiment to analyze the number of manually labeled image pairs needed for a robust estimation of free space. In Fig.~\ref{fig_omap_real_models} we compare the free space map generated by models using 25, 50, and 100 image pairs for training. We see that the segmentation quality of the trained model improves quickly with the addition of more training pairs, giving more accurate free space estimates. We note that our algorithm is sample efficient as even a small number of hand-labeled data is enough to generate accurate maps of real-world environments like a test tank.

\section{Conclusion and Future Work}\label{conclusions}
% 1. Unsupervised adversarial networks with different loss function for sonar image filtering
% 2. Multipath removal 
% 3. Sonar image feature matching (not sure if i should include it...)
% 4. Robust underwater occupancy mapping with state estimations for outdoor scenarios
% 5. Using all information within a range images rather than the first obstacles we met for each azimuth
% 5. Real-time purpose
% 6. Planning in underwater robotics
In this paper, we presented a novel application of cGANs to filter noisy sonar images. Compared to conventional filters, our approach can recognize and filter noise patterns better by distinguishing between obstacles and image artifacts. These results are attainable even with a small training set of raw data and hand-labeled mask pairs.

Using both simulated and real-world data, we showcase the applicability of our method to the downstream task of occupancy mapping, highlighting how our denoising method has a significantly better inference of free and occupied space compared to conventional methods.  

For future directions of this work, we aim to find a solution to account for multipath reflections, something our method cannot disambiguate presently. Apart from autonomous planning and exploration, we aim to study the suitability of our method for feature-based SLAM and 3D object reconstruction using imaging sonar.


\bibliographystyle{IEEEtran-normspace}
\IEEEtriggeratref{16}
\bibliography{main}
\balance
\end{document}